\documentclass{article}

\usepackage{arxiv}

\usepackage[utf8x]{inputenc} 
\usepackage[T1]{fontenc}    
\usepackage{hyperref}       
\usepackage{url}            
\usepackage{booktabs}       
\usepackage{amsfonts}       
\usepackage{nicefrac}       
\usepackage{microtype}      
\usepackage{lipsum}

\usepackage{graphicx}
\usepackage{comment}
\usepackage{amsmath,amssymb,bm} 
\usepackage{xcolor}
\usepackage{float}
\usepackage{hhline}
\usepackage{hyperref}

\usepackage{tikz}
\usetikzlibrary {arrows.meta}
\usetikzlibrary{matrix,fit,positioning}
\usetikzlibrary{bending}
\usetikzlibrary{calc}

\DeclareMathOperator*{\Exp}{Exp}

\def \escalafiguras {0.98}

\newcommand{\Dt}{{\Delta \, t}}
\newcommand{\sigmar}{\sigma_{\mathbf{r}}}
\newcommand{\sigmaR}{\sigma_{\mathrm{\bm{{\theta}}}}}
\newcommand{\sigmav}{\sigma_{\mathbf{v}}}
\newcommand{\sigmaa}{\sigma_{\mathbf{a}}}
\newcommand{\sigmaw}{\sigma_{\mathrm{\bm{{\omega}}}}}
\newcommand{\sigmaAlpha}{\sigma_{\mathrm{\bm{{\alpha}}}}}
\newcommand{\JR}{{\bf J}_{\bf R}^{\bf R}}
\newcommand{\Jw}{{\bf J}_{\bm \omega}^{\bf R}}
\newcommand{\Jalpha}{{\bf J}_{\bm \alpha}^{\bf R}}

\title{High-speed event camera tracking}

\author{
  William Chamorro$^{1,2}$, Juan Andrade-Cetto$^{1}$, and Joan Sol\`a$^{1}$  \\
  \\
$^{1}$ Institut de Rob\`otica i Inform\`atica Industrial, CSIC-UPC\\
LLorens i Artigas 4-6\\
Barcelona - Spain \\
  \texttt{\{wchamorro,cetto,jsola\}@iri.upc.edu} \\
 \\ 
$^{2}$ Universidad UTE \\
Facultad de Ciencias e Ingenier\'ia \\
Quito - Ecuador\\
\texttt{william.chamorro@ute.edu.ec} \\
  
}

\begin{document}
\maketitle

\begin{abstract}
Event cameras are bioinspired sensors with reaction times in the order of microseconds. 
This property makes them appealing for use in highly-dynamic computer vision applications. 
In this work, we explore the limits of this sensing technology and present an ultra-fast tracking algorithm able to estimate six-degree-of-freedom motion with dynamics over 25.8\,g, at a throughput of 10\,kHz, processing over a million events per second. 
Our method is capable of tracking either camera motion or the motion of an object in front of it, using an error-state Kalman filter formulated in a Lie-theoretic sense. 
The method includes a robust mechanism for the matching of events with projected line segments with very fast outlier rejection. 
Meticulous treatment of sparse matrices is applied to achieve real-time performance. 
Different motion models of varying complexity are considered for the sake of comparison and performance analysis.
\end{abstract}


\section{Introduction}

\begin{figure}[t]
\centering
\includegraphics[width=\escalafiguras\linewidth]{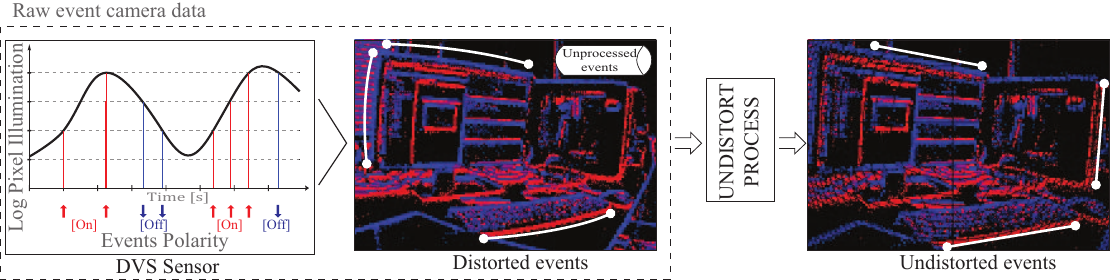}
\caption{Working principle of event cameras (left) with distorted (center) and undistorted output (right).}
\label{fig_distortion}
\end{figure}

Event cameras send independent pixel information as soon as their intensity change exceeds an upper or lower threshold, generating ``ON" or ``OFF" events respectively (see Fig.\ref{fig_distortion}). In contrast to conventional cameras --in which full images are given at a fixed frame rate--, in event cameras, intensity-change messages come asynchronously per pixel, this happening at the microsecond resolution. 
Moreover, event cameras exhibit high dynamic range in luminosity (e.g. 120dB for the Davis 240C model \cite{brandli2014_240} used in this work). 
These two assets make them suitable for applications at high-speed and/or with challenging illumination conditions (low illumination levels or overexposure).
Emerging examples of the use of these cameras in mobile robotics are: event-based optical flow for micro-aerial robotics~\cite{pijnacker2018opFlow_UAV}, obstacle avoidance~\cite{mueggler2015_obst_avoidance,Falanga_scr20}, simultaneous localization and mapping (SLAM)~\cite{Weikersdorfer2014_3d_slam} \cite{milford2015towards_event_slam}, and object recognition~\cite{orchard2015object_recognition}, among others.

We are interested in accurately tracking high-speed 6DoF motion with an event camera. 
This type of sensors has been used in the past for the tracking of motion. 
For instance, 2D position estimates are tracked with the aid of a particle filter in~\cite{weikersdorfer2012event}. 
The method was later extended into an SO(2) SLAM system in which a planar map of the ceiling was reconstructed~\cite{weikersdorfer2013simultaneous}. 
Another SLAM system that tracks only camera rotations and builds a high-resolution spherical mosaic of the scene was presented in~\cite{kim2014_mosaicing}.
Full 3D tracking is proposed in~\cite{kim2016real} where three interleaved probabilistic filters perform pose tracking, scene depth and log intensity estimation as part of a SLAM system. These systems were not designed with high-speed motion estimation in mind.

More related to our approach is the full 3D tracking for high-speed maneuvers of a quadrotor with an event camera presented in~\cite{mueggler2014_track_fast_maneouvers}, extended later to a continuous-time trajectory estimation solution \cite{Mueggler2015}. 
The method is similar to ours in that it localizes the camera with respect to a known wire-frame model of the scene by minimizing point-to-line reprojection errors. 
In that work, the model being tracked is planar, whereas we are able to localize with respect to a 3D model.  
That system was later modified to work with previously built photometric depth maps~\cite{gallego2017event_tracking_from_depthmap}. Non-linear optimization was included in a more recent approach~\cite{bryner2019event_depth_map};in this case, the tracking was performed in a sparse set of reference images, poses and depth maps, by having an a priori initial pose guess and taking into account the event generation model to reduce the number of outliers. 
This event generation model was initially stated in~\cite{gallego2015_generative_tracker} for tracking position and velocity in textured known environments. 
In a more recent contribution, a parallel tracking and mapping system following a geometric, semi-dense approach was presented in~\cite{rebecq2017_event_PTAM}. 
The pose tracker is based on edge-map alignment using inverse compositional Lucas-Kanade method; additionally, the scene depth is estimated without intensity reconstruction. In that work, pose estimates are computed at a rate of 500\,Hz. 

In the long run, we are also interested in developing a full event-based SLAM system with parallel threads for tracking and mapping, that is able to work in real-time on a standard CPU.
Since event cameras naturally respond to edges in the scene, the map, in our case, is made of a set of 3D segments sufficiently scattered and visible to be tracked.
This work deals with the tracking part, and thus such map is assumed given. 
With fast motion applications in mind, our tracking thread is able to produce pose updates in the order of tens of kHz on a standard CPU, 20 times faster than~\cite{rebecq2017_event_PTAM}, is able to process over a million events per second and can track motion direction shifts above 15Hz and accelerations above 25.8\,g.

The main contribution of this paper is first to present a new event-driven Lie-EKF formulation to track the 6DOF pose of a camera in very high dynamic conditions -that runs in real time (10kHz throughput). The use of Lie theory in our EKF implementation allows elegant handling of derivatives and uncertainties in the SO(3) manifold when compared to the classical error-state EKF.
Then we propose a novel fast data association mechanism that robustly matches events to projected 3D line-based landmarks with fast outlier rejection. It reaches real-time performance for over a million of events per second and hundreds of landmarks on a standard CPU.
Finally the benchmarking of several filter formulations including Lie versus classic EKF, three motion models and two projection models, adding up to a total of 12 filter variants.

\section{Motion estimation}\label{section1}
The lines in the map are parametrized by their endpoints ${\bf p}_{\{1,2\}}=(x,y,z)_{\{1,2\}}$ expressed in the object's reference frame. 
We assume the camera is calibrated, and the incoming events are immediately corrected for lens radial distortion using the exact formula in~\cite{Drap2016_undistort_formula}. 

The state vector $\bf x$ represents either the camera state respective to a static object, or the object state respective to a static camera. 
This model duplicity will be pertinent for the preservation of camera integrity in the experimental validation, where tracking very high dynamics will be done by moving the object and not the camera. 

To bootstrap the filter's initial pose, we use the camera's grayscale images.
FAST corners~\cite{rosten2010_FAST_corner} are detected in this 2D image and matched to those in the 3D predefined map. 
The initial pose is then computed using the PnP algorithm~\cite{lepetit2009epnp}. 
After this initial bootstrapping process, the grayscale images are no longer used.
\subsection{Prediction step}
\label{prediction}
The state evolution has the form ${\bf x}_k=f({\bf x}_{k-1},{\bf n}_k)$, where ${\bf n}_k$ is the system Gaussian perturbation. 
The error state $\delta \bf x$ lies in the tangent space of the state, and is modeled as a Gaussian variable with mean $\bar{\delta{\bf x}}$ and covariance $\mathbf{P}$. 

For the sake of performance evaluation, we implemented three different motion models: constant position (CP), constant velocity (CV), and constant acceleration (CA). 
These are detailed in Tab.~\ref{transition}, where ${\bf r}$ represents position, ${\bf R}$ orientation, ${\bf v}$ linear velocity, ${\bm \omega}$ angular velocity, ${\bf a}$ linear acceleration and ${\bm \alpha}$ angular acceleration.
Their Gaussian perturbations are ${\bf r_n},{\bm \theta}_{\bf n},{\bf v_n},{\bm \omega}_{\bf n},{\bf a_n}$, and ${\bm \theta}_{\bf n}$.
%
The orientation $\bf R$ belongs to the $SO(3)$ Lie group, and thus $\bm \omega$ lies in its tangent space $\frak{so}(3)$, although we express it in the Cartesian ${\mathbb R}^3$.
The operator $\oplus$ represents the right plus operation for  $SO(3)$,  ${\bf R} \oplus {\bm\theta} \triangleq {\bf R}\Exp({\bm\theta}) \label{oplus}$, with $\Exp(\cdot)$ the exponential map given by the Rodrigues formula \cite{sola2018micro_lie}.   
\begin{table}[b]
    \centering
        \caption{State transition for {\bf CP}, {\bf CV} and {\bf CA} motion models. Right: error-state partition.}
    \begin{tabular}{r c c | c | c| c || c }
 ${\bf x}_t$ & = & $f( \cdots $ & {\bf CP} & {\bf CV} & ~~~~~~~~~~~~~~{\bf CA} ~~~~~~~~~~~~ ) & $\delta {\bf x}_k $ \\
 \hline
    $\mathbb{R}^3\ni~~{\bf r}_k$ & = & ${\bf r}_{k-1}~+$ & ${\bf r_n}$ & ${\bf v}_{k-1}\Dt$ & ${\bf v}_{k-1}\Dt+\frac12{\bf a}_{k-1}\Dt^2$ & $\delta {\bf r}_k \in \mathbb{R}^3$ \\
    $SO(3)\ni~ {\bf R}_k$ & = & ${\bf R}_{k-1} ~\oplus($ & ${\bm \theta}_{\bf n}$ & ${\bm \omega}_{k-1}\Dt$ & ${\bm \omega}_{k-1}\Dt+\frac12{\bm \alpha}_{k-1}\Dt^2~)$ & $\delta {\bm \theta}_k \in \mathbb{R}^3$ \\
    $\mathbb{R}^3\ni~~{\bf v}_k$ & = & ${\bf v}_{k-1}~+$ && ${\bf v_n}$ & ${\bf a}_{k-1}\Dt$ & $\delta {\bf v}_k \in \mathbb{R}^3$ \\
    $\frak{so}(3)\ni~{\bm \omega}_k$ & = & ${\bm \omega}_{k-1}~+$ && ${\bm \omega}_{\bf n}$ & ${\bm \alpha}_{k-1}\Dt$ & $\delta {\bm \omega}_k \in \mathbb{R}^3$ \\
    $\mathbb{R}^3\ni~~{\bf a}_k$ & = & ${\bf a}_{k-1}~+$ &&& ${\bf a_n}$ & $\delta {\bf a}_k \in \mathbb{R}^3$ \\
    $\mathbb{R}^3\ni~{\bm \alpha}_k$ & = & ${\bm \alpha}_{k-1}~+$ &&& ${\bm \alpha}_{\bf n}$ & $\delta {\bm \alpha}_k \in \mathbb{R}^3$
\end{tabular}
    \label{transition}
\end{table}
%
%

\label{sec:propagation}

The error's covariance propagation is $\mathbf{P}_k = {\bf F}\mathbf{P}_{k-1}{\bf F}^{T}+ \mathbf{Q}\in\mathbb{R}^{m\times m}$ with $m$ equal to $6,12$ or $18$ for the CP, CV, and CA models, respectively. 
$\bf F$ is the Jacobian of $f$ with respect to $\bf x$, and $\mathbf{Q}$ is the perturbation covariance.
Computation for all the motion models is greatly accelerated by exploiting the sparsity of the Jacobian $\bf F$ and the covariance $\bf Q$. We partition these matrices and $\bf P$ in $3\times3$ blocks,
\begin{equation}
\begin{tikzpicture}[
every left delimiter/.style={xshift=.35em},
    every right delimiter/.style={xshift=-.35em},
    node distance=0mm and 0mm,
    baseline]
 \matrix (F) [matrix of nodes,{left delimiter=[},{right delimiter=]}]
{
${\bf I}$ & 0 & ${\bf I}\Delta t$ & 0 & ${\bf I}\Delta t^2$ & 0 \\ 
0 & $\JR$ & 0 & $\Jw$ & 0 & $\Jalpha$ \\ 
0 & 0 & ${\bf I}$ & 0 & ${\bf I}\Delta t$ & 0 \\ 
0 & 0 & 0 & ${\bf I}$ & 0 & ${\bf I}\Delta t$ \\ 
0 & 0 & 0 & 0 & ${\bf I}$ & 0 \\ 
0 & 0 & 0 & 0 & 0 & $~{\bf I}~$ \\ 
};
\node[left=1mm of F]{{\bf F} =};
\node [fit=(F-1-1)(F-2-2), inner sep=-0.5mm, draw, red, thick, name=Fcp] {};
\node [fit=(F-1-1)(F-4-4), inner sep=0mm, draw, teal, thick, name=Fcv] {};
\node [fit=(F-1-1)(F-6-6), inner sep=0.5mm, draw, blue, thick, name=Fca] {};

 \matrix (P)[matrix of nodes,{right= 15mm of F.east},{left delimiter=[},{right delimiter=]}]
{
${\bf P}_{\bf rr}$ & ${\bf P}_{\bf rR}$ & ${\bf P}_{\bf rv}$ & ${\bf P}_{\bf r \bm\omega}$ & ${\bf P}_{\bf ra}$ & ${\bf P}_{\bf r \bm\alpha}$ \\ 
${\bf P}_{\bf Rr}$ & ${\bf P}_{\bf RR}$ & ${\bf P}_{\bf Rv}$ & ${\bf P}_{\bf R \bm\omega}$ & ${\bf P}_{\bf Ra}$ & ${\bf P}_{\bf R\bm\alpha}$  \\ 
${\bf P}_{\bf vr}$ & ${\bf P}_{\bf vR}$ & ${\bf P}_{\bf vv}$ & ${\bf P}_{\bf v\bm\omega}$ & ${\bf P}_{\bf va}$ & ${\bf P}_{{\bf v} {\bm \alpha}}$ \\ 
${\bf P}_{\bm\omega\bf r}$ &${\bf P}_{\bm\omega\bf R}$ & ${\bf P}_{\bm\omega\bf v}$ & ${\bf P}_{\bm\omega\bm\omega}$ & ${\bf P}_{\bm\omega \bf a}$& ${\bf P}_{\bm\omega \bm\alpha}$ \\ 
${\bf P}_{\bf ar}$ & ${\bf P}_{\bf aR}$ & ${\bf P}_{\bf av}$ & ${\bf P}_{\bf a\bm\omega}$ & ${\bf P}_{\bf aa}$ & ${\bf P}_{\bf a \bm \alpha}$ \\ 
${\bf P}_{\bm\alpha\bf r}$ &${\bf P}_{\bm\alpha\bf R}$ & ${\bf P}_{\bm\alpha\bf v}$ & ${\bf P}_{\bm\alpha\bm\omega}$ & ${\bf P}_{\bm\alpha \bf a}$& ${\bf P}_{\bm\alpha \bm\alpha}$ \\ 
};
\node[left=1mm of P]{{\bf P} =};
\node [fit=(P-1-1)(P-2-2), inner sep=-0.5mm, draw, red, thick, name=Pcp] {};
\node [fit=(P-1-1)(P-4-4), inner sep=0mm, draw, teal, thick, name=Pcv] {};
\node [fit=(P-1-1)(P-6-6), inner sep=0.5mm, draw, blue, thick, name=Pca] {};

\node (CP) [draw, red,  thick, text width=0.5cm, below = -13mm of F.east, xshift=7mm, align=center] {CP};
\node (CV) [ draw, teal,  thick, text width=0.5cm,below = 3mm of F.east, xshift=7mm, align=center] {CV};
\node (CA) [ draw, blue,  thick, text width=0.5cm,below = 10mm of F.east, xshift=7mm, align=center] {CA};

\draw[red!60, thick,shorten >=1mm,-{Stealth[bend]}] 
        (CP.west) to [out=180,in=0] (Fcp.east);
\draw[red!60, thick,shorten >=1mm,-{Stealth[bend]}] 
        (CP.east) to [out=0,in=180] (Pcp.west);
\draw[teal!60, thick,shorten >=1mm,-{Stealth[bend]}] 
        (CV.west) to [out=180,in=0] ($(Fcv.east)-(0,8mm)$);
\draw[teal!60, thick,shorten >=1mm,-{Stealth[bend]}] 
        (CV.east) to [out=0,in=180] ($(Pcv.west)-(0,8mm)$);
\draw[blue!60, thick,shorten >=1mm,-{Stealth[bend]}] 
        (CA.west) to [out=180,in=0] ($(Fca.east)-(0,10mm)$);
\draw[blue!60, thick,shorten >=1mm,-{Stealth[bend]}] 
        (CA.east) to [out=0,in=180] ($(Pca.west)-(0,10mm)$);
\end{tikzpicture}
\end{equation}
We follow \cite{sola2018micro_lie} to compute all the non-trivial Jacobian blocks of $\bf F$,
which correspond to the $SO(3)$ manifold. 
Using the notation ${\bf J}^{\bf a}_{\bf b}\triangleq \partial {\bf a}/\partial {\bf b}$, we have
\begin{align}
\label{Jac}
\JR  
= \Exp({\bm\omega} \Dt)^\top 
\;\;\;\text{,}\;\;\;
\Jw 
= {\bf J}_r({\bm\omega} \Dt)\Dt
\;\;\;\text{and}\;\;\;
\Jalpha 
= \tfrac12\Jw\Dt~,
\end{align}
where 
${\bf J}_r(\cdot)$ is the right-Jacobian of $SO(3)$ 
in \cite{sola2018micro_lie}(eq. 143),
%
$\bm{\theta}= {\bm\omega} \Dt \in \mathbb{R}^3$ is a rotation vector calculated as the angular velocity per time, and $[\cdot]_\times\in so(3)$ is a skew symmetric matrix. 
Notice that for CP we have $\JR = \Exp({\bf 0})^\top = \bf I$ and thus ${\bf F}_{CP} = {\bf I}$.

The perturbation covariance $\mathbf{Q}$ is a diagonal matrix formed by the variances ($\sigmar^2$,$\sigmaR^2$) for CP, ($\sigmav^2,\sigmaw^2$) for CV and ($\sigmaa^2,\sigmaAlpha^2$) for CA, times $\Delta t$.
For example, for CV we have ${\bf Q} = \mathrm{block\,diag}({\bf 0}, {\bf 0}, \sigmav^2{\bf I},\sigmaw^2{\bf I})\Dt$.
The $3\times3$ blocks of $\bf P$ are propagated in such a way that trivial operations (add $\bf 0$, multiply by $\bf 0$ or by $\bf I$) are avoided, as well as the redundant computation of the symmetric blocks. 
For example the blocks ${\bf P}_{\bf Rr}$, ${\bf P}_{\bf rR}$ and ${\bf P}_{\bf vv}$ in CV are propagated as,
$\mathbf{P}_{\bf Rr} \gets \JR(\mathbf{P}_{\bf Rr} + \mathbf{P}_{\bf Rv} \Delta t) + \Jw(\mathbf{P}_{ \bm\omega\bf r} + \mathbf{P}_{\bm\omega\bf v} \Delta t)$,~~
$\mathbf{P}_{{\bf rR}} = \mathbf{P}_{{\bf Rr}}^\top$, and~~
$\mathbf{P}_{{\bf vv}} \gets \mathbf{P}_{{\bf vv}} + \sigmav^2{\bf I}\Delta t$\,.
\subsection{Correction step}
\label{correction}

We have investigated the possibilities of either predicting and updating the filter for each event, or collecting a certain number of events in a relatively small window of time. 
Single-event updates are appealing for achieving event-rate throughput, but the amount of information of a single event is so small that this does not pay off. 
Instead, we here collect a number of events in a small window, make a single EKF prediction to the central time $t_0$ of the window, and proceed with updating with every single event as if it had been received at $t_0$. 
This reduces the number of prediction stages greatly and allows us also to perform a more robust data association.

After predicting the state to the center $t_0$ of the window $\Dt$, all visible segments $S_i,~i\in\{1..N\}$ are projected. We consider two projection models: a moving camera in a static world \eqref{u_cam} and a moving object in front of a static camera \eqref{u_ob}, which are used according to the experiments detailed in Sec.~\ref{experiments},
\newcounter{subeq}
\renewcommand{\thesubeq}{\theequation\alph{subeq}}
\newcommand{\newsubeqblock}{\setcounter{subeq}{0}\refstepcounter{equation}}
\newcommand{\mysubeq}{\refstepcounter{subeq}\tag{\thesubeq}}
\begin{align}
 \newsubeqblock 
 \mysubeq  \textrm{ moving camera:~~~} \underline{\bf u}_j &= {\bf K}{\bf R}^\top({\bf p}_j - {\bf r}) \quad j\in\{1,2\}&&\in \mathbb{P}^2 \label{u_cam}\\
  \mysubeq \textrm{ moving object:~~~} \underline{\bf u}_j &= {\bf K}({\bf r}+{\bf R}{\bf p}_j) \quad~~\, j\in\{1,2\} &&\in \mathbb{P}^2 \label{u_ob}\\
\textrm{ projected line:~~~}{\bf l} &= \underline{\bf u}_1 \times \underline{\bf u}_2 = (a, b, c)^\top && \in \mathbb{P}^2 \label{l}
\end{align}
where $\underline{\bf u}_j=(u,v,w)_j^\top$ are the projections of the $i$-th segment's endpoints ${\bf p}_j\in{\mathbb R}^3,~j\in\{1,2\}$, in projective coordinates, and $\bf K$ is the camera intrinsic matrix.
Jacobians are also computed,
\begin{align}
    {\bf J}^{\bf l}_{\bf r} = {\bf J}^{\bf l}_{{\bf u}_i}{\bf J}^{{\bf u}_i}_{\bf r}
    ~~~~~\textrm{and}~~~~~
    {\bf J}^{\bf l}_{\bf R} = {\bf J}^{\bf l}_{{\bf u}_i}{\bf J}^{{\bf u}_i}_{\bf R}
    && \in{\mathbb{R}^{3\times 3}}
    ~,
\end{align}
having  ${\bf J}^{\bf l}_{{\bf u}_1} = -[{\bf u}_2]_\times$, ${\bf J}^{\bf l}_{{\bf u}_2} = [{\bf u}_1]_\times$, ${\bf J}^{{\bf u}_1}_{\bf r} = - {\bf KR}^\top$ for (\ref{u_cam}), ${\bf J}^{{\bf u}_i}_{\bf r}={\bf K}$ for (\ref{u_ob}), and ${\bf J}^{{\bf u}_i}_{\bf R}$ is the Jacobian of the rotation action computed in the Lie-theoretic sense~\cite{sola2018micro_lie}, which for the two projection models becomes
\begin{subequations}
\begin{align}
   \textrm{ moving camera:~~~~~~}{\bf J}^{{\bf u}_i}_{\bf R} &= {\bf K}[{\bf R}^\top({\bf p}_i - {\bf r})]_\times && \in{\mathbb{R}^{3\times 3}}
   \\
   \textrm{ moving object:~~~~~~}{\bf J}^{{\bf u}_i}_{\bf R} &= -{\bf KR}[{\bf p}_i]_\times && \in{\mathbb{R}^{3\times 3}}~.
\end{align}
\end{subequations}

Then, each undistorted event $\mathbf{e}=(u_e, v_e)^\top$ in the window is matched to a single projected segment ${\bf l}$.  
On success (see Sec.~\ref{sec:outlier-rejection} below), we define the event's innovation as the Euclidean distance to the matched segment on the image plane, with a measurement noise $n_d\sim \mathcal{N}(0,\sigma_d^2)$,
\begin{align}
    \textrm{distance innovation :~~~}    z &= d({\bf e}, {\bf l}) = \frac{\underline{\bf e}^\top{\bf l}}{\sqrt{a^2 + b^2}} && \in \mathbb{R}~,\label{z}
\end{align}
where $\underline{\bf e}=(u_e,v_e,1)^\top$. 
The scalar innovation variance is given by ${Z}= {\bf J}_{\bf x}^{z}\mathbf{P}{{\bf J}_{\bf x}^{z}}^\top + \sigma_d^2 \in \mathbb{R}$, where the Jacobian ${\bf J}_{\bf x}^{z}$ of the innovation with respect to the state is a sparse row-vector with zeros in the velocity and acceleration blocks for the larger CV and CA models, 
\begin{align}
	{\bf J}_{\bf x}^{z} &=\begin{bmatrix}{\bf J}_{\bf r}^{z} & {\bf J}_{\bf R}^{z} & {\bf 0} & \cdots & {\bf 0}\end{bmatrix} 
	=
	\begin{bmatrix}{\bf J}^{z}_{\bf l}{\bf J}^{\bf l}_{\bf r} & {\bf J}^{z}_{\bf l}{\bf J}^{\bf l}_{\bf R} & {\bf 0} & \cdots & {\bf 0}\end{bmatrix} 
	&& \in{\mathbb{R}^{1\times m}}~,
\end{align}
with ${\bf J}^{z}_{\bf l} = \underline{\bf e}^\top/\sqrt{a^2 + b^2}$ the Jacobian of (\ref{z}).
At this point an individual compatibility test on the Mahalanobis norm of the innovation is evaluated, $ \frac{z^2}{Z} < n_\sigma^2$, with $n_\sigma \sim 2$.
Upon satisfaction, the Lie-EKF correction is applied:
\begin{equation}\label{eq:ekf_update}
    \begin{aligned}[c]
    \mathrm{a}&)&\;\text{Kalman gain} &:&\; \mathbf{k} &= \mathbf{P}{{\bf J}_{\bf x}^{z}}^\top {Z}^{-1}~~~~~~~  \\
	\mathrm{b}&)&\;\text{Observed error} &:&\;\delta {\bf x} &= \mathbf{k}{z}
    \end{aligned}
    \;\;
    \begin{aligned}[c]
    \mathrm{c}&)&\;\text{State update} &:&\;{\bf x} &\gets {\bf x} \oplus \delta {\bf x}\\
	\mathrm{d}&)&\;\text{Cov. update} &:&\;\mathbf{P} &\gets \mathbf{P}-\mathbf{k}{Z}\mathbf{k}^\top~,
    \end{aligned}
\end{equation}{}%
where the state update $c)$ is implemented by a regular sum for the state blocks $\{{\bf r}, {\bf v}, {\bm\omega}, {\bm\alpha}\}$ and by the right-plus ${\bf R}\Exp(\delta{\bf \theta})$ for ${\bf R}\in SO(3)$, as needed for the model in turn (CP, CV, or CA).
We remark for implementation purposes affecting execution speed that the Kalman gain $\bf k$ is an $m-$vector, that to compute $Z$ and (\ref{eq:ekf_update}a) we again exploit the sparsity of ${\bf J}^{z}_{\bf x}$, as we did in~\ref{sec:propagation}, and that $Z^{-1}$ is the inverse of a scalar.
\subsection{Fast event-to-line matching}
\label{sec:outlier-rejection}
\begin{figure}[t]
\centering
\includegraphics[width=\escalafiguras\linewidth]{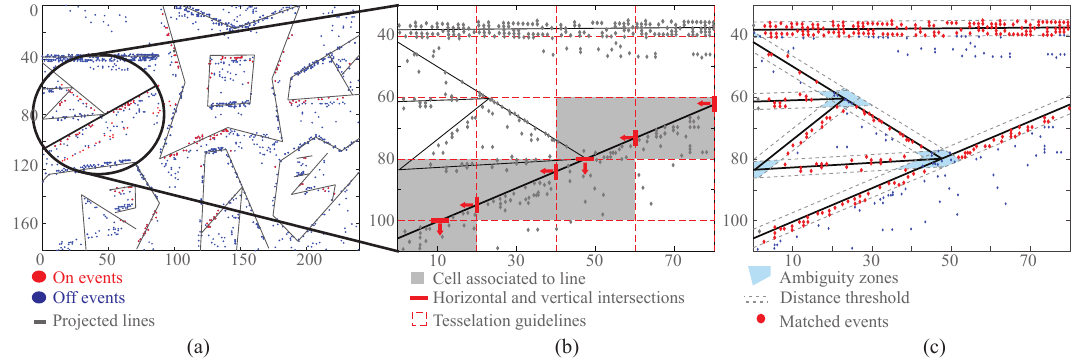}
\caption{Data association process: (a) event window sample with projected lines, (b) cell identification for a single line based on the tessellation guidelines, (c) thresholding and ambiguity removal.}  
\label{fig_window_events}
\end{figure}

The Lie-EKF update described above is preceded by event outlier detection and rejection.
The goal is to discard or validate events rapidly before proceeding to the update.
We use image tessellation to accelerate the search for event-to-line candidates.
To do so, for each temporal window of events, we first identify the visible segments in the map and project them
using either (\ref{u_cam}) or (\ref{u_ob}) as appropriate. Fig.~\ref{fig_window_events}(a) displays a capture of a temporal window of events of 100$\mu s$. 

The image is tessellated in $m\times n$ (reasonably squared) cells, each one having a list of the segments crossing it. 
These lists are re-initialized at the arrival of each new window of events. 
The cells ($C_u,C_v$) crossed by a segment are identified by computing the segment intersections with the horizontal and vertical tessellation grid lines.
We use the initial segment endpoint coordinates ($u_0,v_0$), the line parameters (\ref{l}) and the image size $w \times h$, see Fig~\ref{fig_window_events}(b). 
The computation of the crossed cells coordinates departs from a cell given by the initial segment endpoint, $C_u^{0} = \lceil u_0 m/w \rceil$ and $C_v^{0} = \lceil(v_0 n/h \rceil$, where  $\lceil C \rceil \triangleq \mathrm{ceil}(C)$.
Then we sequentially identify all horizontal and vertical intersections,
%
\begin{align}
    \text{Horiz: }& 
   \begin{cases}
    C^{i}_v =  i \\
    C^i_u = \lceil(-b h i -c n )m/a n w  \rceil
     \end{cases}
    & \text{Vert: }&
     \begin{cases}
     C^{j}_u =  j\\
     C^{j}_v = \lceil(-aw j-c m )n/b m h  \rceil,
    \end{cases}
\end{align}
where the iterators $i$ and $j$ keep track of the horizontal and vertical intersections. Their values start from $C_v^{0}$ and $C_u^{0}$, 
respectively and are increased or decreased by one in each iteration until reaching the opposite endpoint cell location. The sign of the increment depends on the difference between the first and last endpoint cell coordinates. 

For each event in the temporal window we must check whether it has a corresponding line match in its corresponding tessellation cell.
Although we are capable of processing all events at the rate of millions per second, there might be cases in which this is not achievable due to a sudden surge of incoming events. 
This depends on the motion model used, the scene complexity, or the motion dynamics. We might need to leave out up to 1/10th of events on average in the most demanding conditions (see last row of Tab.~\ref{table_metrics}), and to do so unbiasedly, we keep track of execution time and skip the event if its timestamp is lagging more than 1$\mu s$ from the current time.

Each unskipped event inside each cell is compared only against the segments that are within that cell. 
This greatly reduces the combinatorial explosion of comparing $N$ segments with a huge number $M$ of events from $O(M\times N)$ to the smaller cost of updating the cells' segments lists, which is only $O(N)$.
The (very small) number of match segment candidates for each event are sorted from min to max distance. 
To validate a match between an event and its closest segment the following three conditions (evaluated in this order) must be met, see Fig~\ref{fig_window_events}(c): 
a) the distance $d_1$ (\ref{z}) to the closest segment is \textit{below} a predefined threshold, $d_1 < \alpha$;
b) the distance $d_2$ to the second closest segment is \textit{above} another predefined threshold, 
$d_2 > \beta$; and
c) the orthogonal projection of the event onto the segment falls between the two endpoints, $0 < \frac{{\bf v}_1^\top{\bf v}_2}{{\bf v}_1^\top{\bf v}_1}< 1 $, where ${\bf v}_1 = {\bf u}_2 - {\bf u}_1$,  ${\bf v}_2 = {\bf e} - {\bf u}_1$, and ${\bf u}_i$ are the endpoints in pixel coordinates. 
Events that pass all conditions are used for EKF update as described in Sec.~\ref{correction}.

\section{Experiments and results}\label{experiments}

Our algorithm is set up with a temporal event-window size of $\Dt=100 \mu s$.
The continuous-time perturbation parameters of the motion models (see Sec.~\ref{prediction}) are listed in  Tab.~\ref{table_perturbation_parameters}, and the outlier rejection thresholds are set at $\alpha=2.5$ pixels, and $\beta=3.5$ pixels.
These parameters were set in accordance with the velocities and accelerations expected. 
The experiments were carried out with different random camera hand-movements and a four-bar mechanism to test the tracking limits of our approach. For future comparisons, the dataset used to generate this results, parametrized maps using endpoints, camera calibration parameters, and detailed information about data format is available at \url{https://www.iri.upc.edu/people/wchamorro/}.

We make use of our C++ header-only library  ``{\texttt{manif}}''~\cite{deray2019_manif} for ease of Lie theory computations. 
The event-rate tracker runs single-threaded on standard PC hardware with Ubuntu 16.04.5 LTS and ROS Kinetic.
\noindent
\begin{table}[t]
\centering
\makebox[0pt][c]{\parbox{\escalafiguras\textwidth}{%
    \begin{minipage}[t]{\dimexpr.25\textwidth-.5\columnsep}
    \smallskip\noindent
    \centering
        \caption{Perturbation and noise parameters.}
        \label{table_perturbation_parameters}
        \footnotesize
        \begin{tabular}{c|cl}
            \hline
            ${\bm \sigma}$ & \multicolumn{2}{c}{\bf Value}\\
            \hline
            $\sigmar$ & 0.03 & m/s$^{1/2}$\\
            $\sigmaR$ & 0.3  & rad/s$^{1/2}$\\
            $\sigmav$ &  3   & m/s$^{3/2}$\\
            $\sigmaw$ &  10  & rad/s$^{3/2}$\\
            $\sigmaa$ &  80  & m/s$^{5/2}$\\
            $\sigmaAlpha$ & 300 & rad/s$^{5/2}$\\
            \hline
            $\sigma_d$ & 3.5 & pixels\\ 
            \hline
        \end{tabular}
    \end{minipage}
    \hfill
    \begin{minipage}[t]{\dimexpr.7\textwidth-.5\columnsep}
    \setlength{\tabcolsep}{4pt}
    \smallskip\noindent
    \centering
        \caption{RMSE mean values and timings. {\bf L}: Lie parameterization, and {\bf Cl}: classic algebra.}
        \label{table_metrics}
        \footnotesize
        \begin{tabular}{c|cccccc}
        \hline
        Metric & {\bf CP}+{\bf L} & {\bf CV}+{\bf L} & {\bf CA}+{\bf L} & {\bf CP}+{\bf Cl} & {\bf CV}+{\bf Cl} & {\bf CA}+{\bf Cl}\\
        \hline
        $x$ (m)  & 0.0149 & 0.0091 & 0.0095 & 0.0162 & 0.0093 & 0.0106\\
        $y$ (m)  &0.0125 & 0.0085 & 0.0081 &0.0119 & 0.0086 & 0.0088\\
        $z$ (m)  & 0.0167& 0.0111 &0.0012 & 0.0171 & 0.0121 & 0.0113\\
        $\phi$ (rad)  & 1.2205 & 0.7522 & 0.8333 & 1.2729 & 0.8613 & 0.9539\\
        $\theta$ (rad) & 1.4569 & 0.9842 & 1.0209 &1.2729 & 1.2366 & 1.2645\\
        $\psi$ (rad) & 1.2955 & 0.9252 & 0.8066 &1.1549 & 1.1201 & 0.9902\\
        \hline
        $T_{proc}$ ($\mu s$) & 0.32 & 0.46 & 0.72 &0.29 & 0.42 & 0.64\\
        $N_{events}$ ($\%$) & 97.73 & 90.96 & 85.51 & 98.06 & 92.68 & 89.09\\ 
        \hline
        \end{tabular}
    \end{minipage}
}}
\end{table}
\subsection{Position and orientation RMSE evaluation}
\label{RMS_evaluation}
The RMSE evaluation allow us to statistically determine the accuracy and consistency of the tracker. 
We execute $N=10$ Monte Carlo runs of different motion sequences of about 60\,s of duration in different conditions such as: speed changes from low (0.5\,m/s, 3\,rad/s approx. avg.) to fast (1\,m/s, 8\,rad/s approx. avg.), moving the camera manually inside the scene in random trajectories; aleatory lightning changes, turning on/off the laboratory lights and strong rotation changes.
\begin{figure}[b]
\centering
\includegraphics[width=\escalafiguras\linewidth]{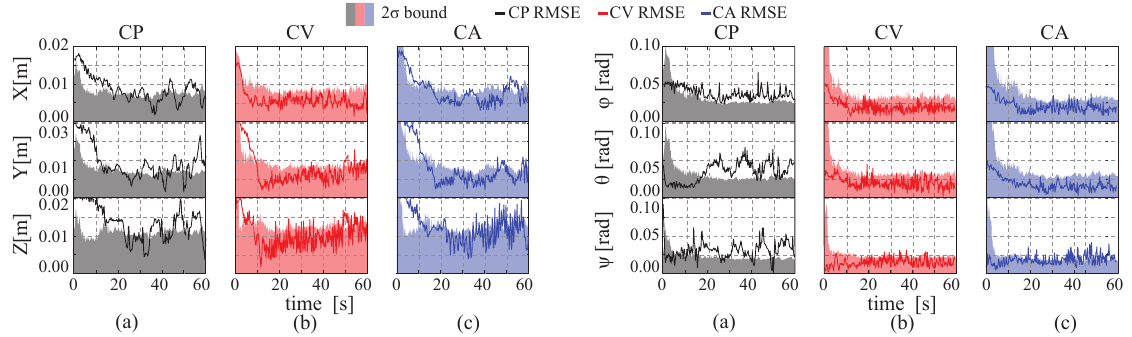}
\caption{RMS errors and 2-sigma bounds: (a-c) position, (d-f) orientation.}
\label{fig_errors}
\end{figure}
In this evaluation, we use the projection model (\ref{u_cam}); i.e., the camera is moving in a static world. 
From the 10 runs, we measure the root mean square error (RMSE) of each component of the camera pose  and plot it in Fig.~\ref{fig_errors}. To analyze consistency of the filter, the errors obtained are compared against their 2-sigma bounds as in \cite{Sola2011_impact_lanmark_parametrization}. 
An OptiTrack motion capture system calibrated with spherical reflective references will provide the ground truth to analyze the event-based tracker performance.
For the sake of comparison, we also implemented the classic ES-EKF using quaternions, where Jacobians are obtained using first-order approximations. 
The error evaluation for the various filter variants tested are summarized in Tab.~\ref{table_metrics}.
\begin{figure}[t!]
\centering
\includegraphics[width=\escalafiguras\linewidth]{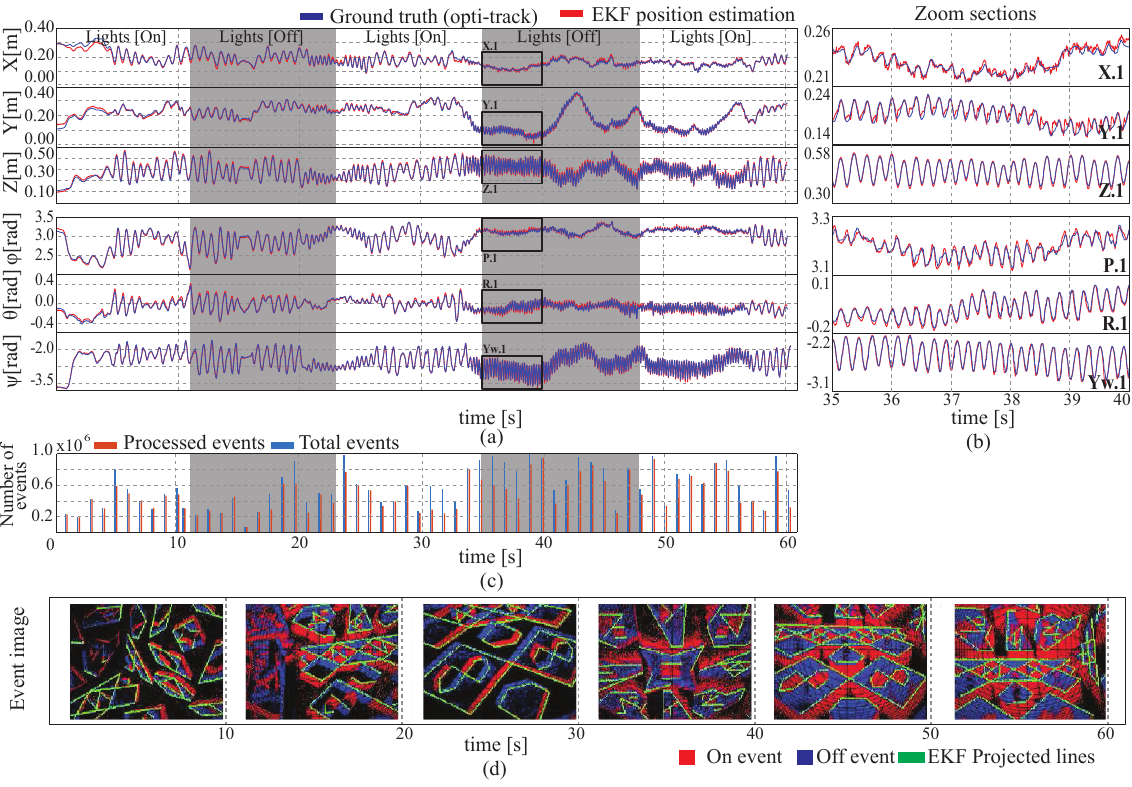}
\caption{(a) Strong hand shake ($\sim$ 6Hz) sequence example (using {\bf CV+L}), (b) with zoom in the high speed zone, (c) event quantification and (d) visual output snapshots at a given time.}
\label{fig_pose}
\end{figure}

The overall results show a small but noticeable improvement in accuracy when the tracker is implemented with Lie groups, where the CV model has the best response.
Though the Lie approach is somewhat slower, this can be taken as the price to pay for improved accuracy.
During the RMSE evaluation, CV and CA errors were mostly under the 2-sigma bound (see Fig.~\ref{fig_errors} (b,c,e,f)) indicating a sign of consistency. 
On the other hand, the error using the CP model is shown to exceed the 2-sigma bound repeatedly. 
This situation was evidenced during the experiments by observing less resilience of the tracker in high dynamics (see Fig.~\ref{fig_errors} (a,d)).

In all cases, per-event total processing time $T_{proc}$ falls well bellow the microsecond, where, on average, less than $0.1\,\mu$s of this time is spent performing line-event matching, the rest being spent in prediction and correction operations.
With this, the tracker is capable of treating between 89.1\% and 97.7\% of the incoming data, depending on the motion model and state parameterization used, reaching real-time performance, and producing pose updates at the rate of 10\,kHz, limited only by the chosen size of the time window of events of 100\,$\mu$s.

A comparison of the tracker performance versus the OptiTrack ground truth is shown in Fig.~\ref{fig_pose} for the best performing motion model and state parameterization combination: constant velocity with Lie groups.
In this case, the camera is hand-shaked by a human in front of the scene.
The frequency of the motion signal increases from about 1\,Hz to 6\,Hz, the fastest achievable with a human hand-shake of the camera.
The camera pose is accurately tracked despite the sudden changes in motion direction, where the most significant errors  --in the order of mm-- are observed precisely in these zones where motion changes direction (see Fig~\ref{fig_pose}(a,b)).
Illumination changes were produced by turning on and off the lights in the laboratory with no noticeable performance degradation in the tracking nor the event production (see grey shaded sections in Figs.~\ref{fig_pose}(a),(c)), which reached peaks of about one million events per second with the most aggressive motion dynamics (see zoomed-in region in Fig.~\ref{fig_pose}(b) and (c)).
The green lines in the snapshots in Fig.~\ref{fig_pose}(d) are the projected map segments using the estimated camera pose.
\subsection{High speed tracking}
\begin{figure}[t!]
\centering
\includegraphics[width=\escalafiguras\linewidth]{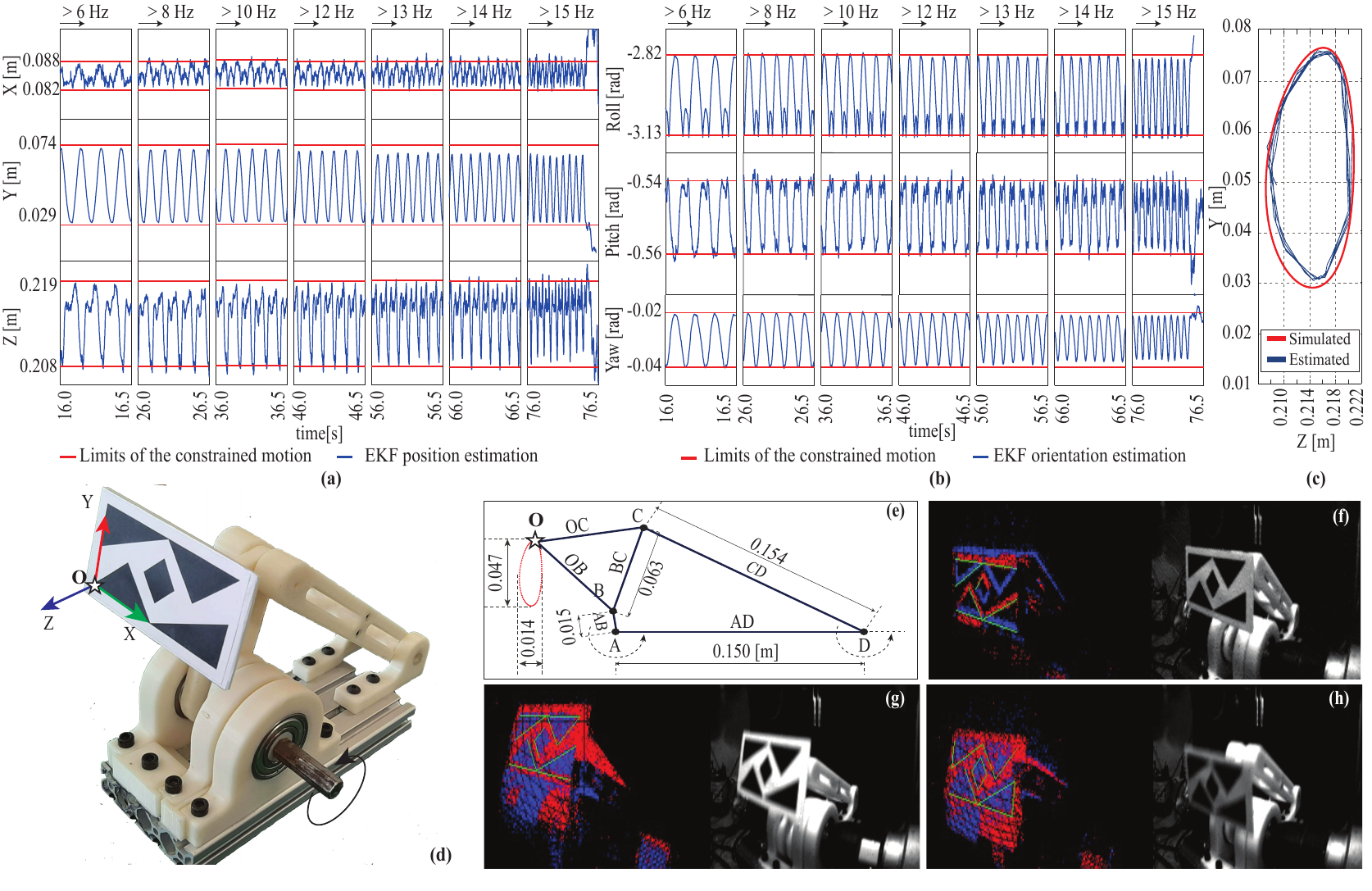}
\caption{High dynamics position and orientation evaluation using $\bf CV+L$: (a,b), poses up to 950\,rpm (15.8\,Hz) were accurately estimated before the tracking disengaged, (c) Z-Y trajectory (d) constrained four-bar motion mechanism, (e) mechanism dimensions and (f-h) visual snapshots of the tracker for crank angular speeds of 300, 500 and 800\,rpm respectively.}
\label{fig_freq}
\end{figure}
Our aim is to explore the limits of this sensing technology and submit the event camera to the highest dynamics it is able to track, and provide a mean of comparison with other approaches in terms of speed. 
To protect the camera from destructive vibrations, we switch now to a motion model in which the camera is stationary and the tracked object moves in front of it, using Eq.~\ref{u_ob} for the projections. 
This experiment is reported for our best-achieving version of the tracker, CV with Lie parametrization.

We built a constrained motion device (see Fig.~\ref{fig_freq}(d)), consisting of a four-bar mechanism powered with a DC motor and dimensions stated in Fig.~\ref{fig_freq}(e). The mechanism very-rapidly shakes a target made of geometric shapes, delimited by straight segments, in front of the camera.

A kinematic analysis performed for our mechanism using real dimensions gives a 4.7\,cm maximum displacement of the target reference frame (peak-to-peak). An evaluation point in the target (e.g. $\bm O$ in Fig.~\ref{fig_freq}(d)) describes a motion with rotational and translational components, and has a simulated trajectory as the one in red dots in Fig.~\ref{fig_freq}(e) (around the axes Y and Z).

The estimated trajectory of the object in the axes Y and Z was compared to the simulated one considering the previously known dimensions and constrains of the mechanism. 
The estimated trajectory is within the motion limits and has a low associated error as can be seen in Fig.~\ref{fig_freq}(c). For visualization purposes we plotted ten motion periods chosen randomly along the running time.
The camera is placed statically at a distance of roughly 20\,cm in front of the target while a DC motor drives the mechanism.
During the experiment (see Fig.~\ref{fig_freq}(a,b)) its speed was increased gradually until about 950\,rpm (15.8\,Hz) where tracking performance starts to degrade. At such crank angular velocity, the velocity analysis of our mechanism reports a maximum target speed of 2.59\, m/s.

Linear target accelerations reach over 253.23\,m/s$^2$ or 25.81\,g, which are well above the expected range of the most demanding robotics applications, and above the maximum range of 16\,g of the user-programmable IMU chip in the Davis 240c camera~\cite{muggler2017event_phd}.  The green lines in Fig.\ref{fig_freq}(f-h) are plot using the estimated camera pose.

\section{Conclusions and future work}
In this paper, an event-based 6-DoF pose estimation system is presented. Pose updates are produced at a rate of 10 kHz with an error-state Kalman filter.
Several motion model variants are evaluated, and it is shown that the best performing motion model and pose parameterization combination is a constant velocity model with Lie-based parameterization.  
In order to deal with the characteristic \emph{micro second} rate of event cameras, a very fast event-to-line association mechanism was implemented. 
Our filter is able to process over a million events per second, making it capable of tracking very high-speed camera or object motions without delay.
Considering the low resolution of the camera, our system is able to track its position with high accuracy, in the order of a few mm with respect to a calibrated OptiTrack ground truth positioning system.
Moreover, when subjected to extreme motion dynamics, the tracker was able to reach tracking performance for motions exceeding linear speeds of 2.5\,m/s and accelerations over 25.8\,g.
Our future work will deal with the integration of this localization module into a full parallel tracking and mapping system based on events.
\subsection*{Acknowledgements} 
This work was partially supported by the EU H2020 project GAUSS (H2020-Galileo-2017-1-776293), by the Spanish State Research Agency through projects EB-SLAM (DPI2017-89564-P) and the Mar\'ia de Maeztu Seal of Excellence to IRI (MDM-2016-0656, and by a scholarship from SENESCYT, Republic of Ecuador to William Chamorro.%

\bibliographystyle{unsrt}  


\begin{thebibliography}{10}

\bibitem{brandli2014_240}
Christian Brandli, Raphael Berner, Minhao Yang, Shih~Chii Liu, and Tobi
  Delbruck.
\newblock {A 240 × 180 130 dB 3 $\mu$s latency global shutter spatiotemporal
  vision sensor}.
\newblock {\em IEEE J.~Solid-State Circuits}, 49(10):2333--2341, 2014.

\bibitem{pijnacker2018opFlow_UAV}
Bas~J. {Pijnacker Hordijk}, Kirk~Y.W. Scheper, and Guido~C.H.E. de~Croon.
\newblock {Vertical landing for micro air vehicles using event-based optical
  flow}.
\newblock {\em J.~Field Robotics}, 35(1):69--90, 2018.

\bibitem{mueggler2015_obst_avoidance}
Elias Mueggler, Nathan Baumli, Flavio Fontana, and Davide Scaramuzza.
\newblock {Towards evasive maneuvers with quadrotors using dynamic vision
  sensors}.
\newblock In {\em Eur. Conf. Mobile Robots}, pages 1--8, 2015.

\bibitem{Falanga_scr20}
Davide~Scaramuzza Davide~Falanga, Kevin~Klever.
\newblock Dynamic obstacle avoidance for quadrotors with event cameras.
\newblock {\em Sci. Robotics}, 5(40):eaaz9712, 2020.

\bibitem{Weikersdorfer2014_3d_slam}
David Weikersdorfer, David Adrian, Daniel Cremers, and Jorg Conradt.
\newblock {Event-based 3D SLAM with a depth-augmented dynamic vision sensor}.
\newblock In {\em IEEE Int. Conf. Robotics Autom.}, pages 359--364, 2014.

\bibitem{milford2015towards_event_slam}
Michael Milford, Hanme Kim, Stefan Leutenegger, and Andrew Davison.
\newblock Towards visual {SLAM} with event-based cameras.
\newblock {\em {RSS} Workshop on the Problem of Mobile Sensors}, 2015.

\bibitem{orchard2015object_recognition}
Garrick Orchard, Cedric Meyer, Ralph Etienne-Cummings, Christoph Posch, Nitish
  Thakor, and Ryad Benosman.
\newblock {HFirst: A temporal approach to object recognition}.
\newblock {\em IEEE Trans. Pattern Anal. Mach. Intell.}, 37(10):2028--2040,
  2015.

\bibitem{weikersdorfer2012event}
David Weikersdorfer and Jorg Conradt.
\newblock {Event-based particle filtering for robot self-localization}.
\newblock In {\em IEEE Int. Conf. Robotics Biomim.}, pages 866--870, 2012.

\bibitem{weikersdorfer2013simultaneous}
David Weikersdorfer, Raoul Hoffmann, and J{\"{o}}rg Conradt.
\newblock {Simultaneous localization and mapping for event-based vision
  systems}.
\newblock In {\em Int. Conf. Comput. Vis. Syst.}, pages 133--142, 2013.

\bibitem{kim2014_mosaicing}
Hanme Kim, Ankur Handa, Ryad Benosman, Sio-Hoi Ieng, and Andrew~J Davison.
\newblock Simultaneous mosaicing and tracking with an event camera.
\newblock {\em IEEE Journal of Solid-State Circuits}, 43:566--576, 2008.

\bibitem{kim2016real}
Hanme Kim, Stefan Leutenegger, and Andrew~J Davison.
\newblock {Real-time 3D reconstruction and 6-DoF tracking with an event
  camera}.
\newblock In {\em Eur. Conf. Comput. Vis.}, pages 349--364, 2016.

\bibitem{mueggler2014_track_fast_maneouvers}
Elias Mueggler, Basil Huber, and Davide Scaramuzza.
\newblock {Event-based, 6-DOF pose tracking for high-speed maneuvers}.
\newblock In {\em IEEE/RSJ Int. Conf. Intell. Robots Syst.}, pages 2761--2768,
  2014.

\bibitem{Mueggler2015}
Elias Mueggler, Guillermo Gallego, and Davide Scaramuzza.
\newblock Continuous-time trajectory estimation for event-based vision sensors.
\newblock In {\em Robotics Sci. Syst. Conf.}, 2015.

\bibitem{gallego2017event_tracking_from_depthmap}
Guillermo Gallego, Jon~E.A. Lund, Elias Mueggler, Henri Rebecq, Tobi Delbruck,
  and Davide Scaramuzza.
\newblock Event-based {6-DOF} camera tracking from photometric depth maps.
\newblock {\em pami}, 40(10):2402--2412, 2017.

\bibitem{bryner2019event_depth_map}
Samuel Bryner, Guillermo Gallego, Henri Rebecq, and Davide Scaramuzza.
\newblock Event-based direct camera tracking from a photometric {3D} map using
  nonlinear optimization.
\newblock In {\em IEEE Int. Conf. Robotics Autom.}, pages 325--331, 2019.

\bibitem{gallego2015_generative_tracker}
Guillermo Gallego, Christian Forster, Elias Mueggler, and Davide Scaramuzza.
\newblock Event-based camera pose tracking using a generative event model.
\newblock {\em arXiv: 1510.01972}, 1:1--7, 2015.

\bibitem{rebecq2017_event_PTAM}
Henri Rebecq, Timo Horstschaefer, Guillermo Gallego, and Davide Scaramuzza.
\newblock {EVO: A geometric approach to event-based 6-DOF parallel tracking and
  mapping in real time}.
\newblock {\em IEEE Robotics Autom. Lett.}, 2(2):593--600, 2017.

\bibitem{Drap2016_undistort_formula}
Pierre Drap and Julien Lef{\`{e}}vre.
\newblock {An exact formula for calculating inverse radial lens distortions}.
\newblock {\em Sensors}, 16(6):807, 2016.

\bibitem{rosten2010_FAST_corner}
Rosten Edward, Porter Reid, and Drummond Tom.
\newblock {Faster and better: A machine learning approach to corner detection}.
\newblock {\em IEEE Trans. Pattern Anal. Mach. Intell.}, 32(1):105--119, 2010.

\bibitem{lepetit2009epnp}
Pascal {Lepetit, Vincent and Moreno-Noguer, Francesc and Fua}.
\newblock {EPnP}: An accurate {O(n)} solution to the {PnP} problem.
\newblock {\em Int. J.~Comput. Vision}, 81:155--166, 2009.

\bibitem{sola2018micro_lie}
Joan Sol{\`{a}}, Jeremie Deray, and Dinesh Atchuthan.
\newblock {A micro Lie theory for state estimation in robotics}.
\newblock {\em arXiv: 1812.01537}, pages 1--16, 2018.

\bibitem{deray2019_manif}
Jeremie Deray and Joan Sol{\`{a}}.
\newblock {manif: a small C++ header-only library for Lie theory.}
\newblock https://github.com/artivis/manif, jan 2019.

\bibitem{Sola2011_impact_lanmark_parametrization}
Joan Sol{\`{a}}, Teresa Vidal-Calleja, Javier Civera, and Jose~Maria
  Martinez-Montiel.
\newblock {Impact of landmark parametrization on monocular EKF-SLAM with points
  and lines}.
\newblock {\em Int. J.~Comput. Vision}, 97:339--368, 2011.

\bibitem{muggler2017event_phd}
Elias Mueggler.
\newblock {\em Event-based Vision for High-Speed Robotics}.
\newblock PhD thesis, University of Zurich, 2017.

\end{thebibliography}

\end{document}